# CLASSify: A Web-Based Tool for Machine Learning

**Aaron D. Mullen, B. S.[1], Samuel E. Armstrong, M. S.[1], Jeff Talbert, Ph. D.[1], V.K. Cody Bumgardner, Ph. D.[1],**
[1]**University of Kentucky, Lexington, KY, USA**

**Abstract**

*Machine learning classification problems are widespread in bioinformatics, but the technical knowledge required to perform model training, optimization, and inference can prevent researchers from utilizing this technology. This article presents an automated tool for machine learning classification problems to simplify the process of training models and producing results while providing informative visualizations and insights into the data. This tool supports both binary and multiclass classification problems, and it provides access to a variety of models and methods. Synthetic data can be generated within the interface to fill missing values, balance class labels, or generate entirely new datasets. It also provides support for feature evaluation and generates explainability scores to indicate which features influence the output the most. We present CLASSify, an open-source tool for simplifying the user experience of solving classification problems without the need for knowledge of machine learning.*

**Introduction**

Classification problems are one of the most common types of problems in bioinformatics[1,2]. For example, patient data may need to be analyzed to determine the presence of a particular disease, or researchers may want to use information about cell or tissue samples to identify medical conditions. Being able to accurately classify patient data is crucial to ensuring they receive the proper diagnosis and treatment. However, manually reviewing large amounts of data to identify trends and compare patients is infeasible for larger datasets. Therefore, machine learning (ML) classification methods are often used for these problems[3].

The process of building and training an ML classification model can be overwhelming for medical professionals without the technical expertise needed to create an effective model, due to the "black box" nature of many classification models that makes them difficult to interpret on their own. Additionally, parameter tuning is difficult to understand and can have a large impact on model results. CLASSify (Classification Learning Automated Software System) solves these problems by providing an easy-to-use interface with options for automated feature evaluation and parameter tuning that provides understandable results detailing the performance of selected models as well as additional insights into the data.

In addition to providing models and performance metrics, CLASSify provides synthetic data generation. The ability to generate realistic synthetic data is imperative in the medical field, as real data may be protected or imbalanced[4,5]. Many diagnosis problems suffer from imbalanced class labels, where most patient or sample data consist of instances without the observed condition. These types of datasets can be difficult for machine learning methods to accurately learn and predict, so synthetic data can be used to introduce additional records that balance class labels and allow the model to better understand the trends of the data.

CLASSify also provides results related to the importance of each feature to the models. These are described through the Shapley Additive Explanations (SHAP) scores[6], which provide values that represent how much each feature of a dataset contributes to a certain prediction. Overall SHAP scores for the model can be obtained by averaging together the SHAP scores for each row of data. These values are useful for gaining a better understanding of the data and the relationships between variables that may not be obvious. For example, a high positive score indicates that the feature contributes positively to the predictions, meaning that higher values of that feature influence the model to classify "1" instead of "0" in a binary classification problem. A high negative score indicates the opposite, where higher feature values lead to predictions closer to 0 than 1.

Finally, CLASSify creates several visualizations for an easier understanding of the results of the model and the importance of each feature.

Other solutions have implemented machine learning models with a web interface to improve the ease with which users can generate models and interpret results[7,8]. The work of Carney et al.[7], called Teachable Machine, is the most similar, but there are a few key differences. For one, Teachable Machine is meant for training and classifying images and sounds, while CLASSify only works with tabular data. All information about the dataset is taken from the uploaded CSV file rather than inputted manually by the user. Additionally, Teachable Machine was built for education, while CLASSify is meant for use in clinical settings. In general, other interfaces have not implemented the robust support for synthetic data generation, SHAP scores, or interpretable results that CLASSify provides, and many are commercial solutions that come with a significant price tag.

**Methods**

CLASSify employs various well-established data processing techniques while integrating bespoke components to enhance user-friendliness for non-specialists. Specifically, a web interface provides simple data ingestion and informative visualizations, which are available for download. In addition to visualizations, results provided by each model are available for closer inspection using an explainability algorithm (namely SHAP). Many machine learning applications may lack an explainability option, a critical aspect, especially given the prevalence of complex black box models presently available.

CLASSify provides support for both binary and multiclass classification. The ClearML[9] platform is used to host the models and distribute the training and predicting tasks efficiently. ClearML also stores datasets and visualization plots generated by CLASSify.

Most models were implemented using the sklearn[10] Python library. For binary classification, supported models include Random Forest[11], Gradient Boosting[12], XGBoost[13], Histogram-based Gradient Boosting[12], Bagging Classifier[14], Multi-Layer Perceptron[15], TabPFN[16], SGD Classifier[17], Logistic Regression[18], and K-Nearest Neighbors[19]. The XGBoost and TabPFN models were provided by the XGBoost[20] and TabPFN[21] libraries respectively. Multiclassification works with Random Forest, Multi-Layer Perceptron, Logistic Regression, and K-Nearest Neighbors. When training with CLASSify, any combination of these provided models can be chosen and results and comparisons between all of them are produced.

Users can also choose to perform feature evaluation on any model, which tests various combinations of features to find which produces the best performance. Users can define which features to use and how many features to include in each iteration. With large numbers of features, testing every possible combination of features may become infeasible to perform in a reasonable amount of time. However, SHAP scores can be provided at the end for each feature regardless of whether the user chooses to test different feature combinations.

For training, users can provide a separate test set, and if they do not, the test set will be split from the dataset provided. These are passed to each model for training and testing.

CLASSify also automatically performs parameter optimization with Optuna[22]. Default parameter ranges provided to each model, such as the number of trees for a Random Forest or the size of the hidden layer for the Multi-Layer Perceptron, are available. Users can manually input these parameter ranges for more control over the model, as different ranges will be suitable for different datasets. Optuna will then optimize the model by trying different parameter combinations within those ranges, intelligently learning which settings give the best results, and pruning unpromising parameter choices. It runs for the specified number of iterations and returns the parameters that performed the best on a validation set. Increasing the number of iterations is likely to always result in better performance at the expense of time. Therefore, users have the choice of whether to value time or performance more heavily.

After the highest-performing iteration for each model is chosen, the tuned model is evaluated on the unseen test set, and the results are stored. Once each model has produced results, the table of performance metrics, which includes accuracy, AUC, sensitivity, specificity, negative predictive value (NPV), and positive predictive value (PPV), is outputted to the user. For multiclassification, the kappa score is provided instead of sensitivity, specificity, NPV, and PPV.

Synthetic data generation is also included and has three versions: fill in missing values, balance class labels, or generate entirely new datasets. These synthetic methods require JSON metadata describing the datatype of each feature. If the user does not provide this metadata themselves, it will be automatically generated.

The method to synthetically fill missing values is the Soft Impute[23] function, provided by the fancyimpute[24] library. This method uses soft-threshold and single-value decomposition (SVD) to remove noise and create new data that follows the trends of the original data. This means that missing data will be replaced with new data that follows closely to the format and pattern that would be expected.

The processes used to generate entire new rows of data are provided by the Synthetic Data Vault (SDV)[25] package. This repository contains several different models for data generation, including Fast ML Preset[26], CTGAN[27], CopulaGAN[27], and TVAE[27] synthesizers. All models have been incorporated into CLASSify, and the user can choose which model is used for data generation.

Both balancing class labels and generating new datasets are slight modifications of the same process. Both use the given SDV model to generate new rows of data following the trends and conditions set by the rest of the data. If the user chooses to generate new data, a new dataset of the same size as the provided dataset is created, but with all class labels automatically balanced. If the user only wants to bolster the existing data by balancing the class labels, the system will simply generate the number of rows for each class as necessary to ensure a balanced dataset and append those synthetic rows onto the original dataset. Therefore, the majority class label will maintain the same number of rows, and any other class labels will be increased to match that value, resulting in a larger dataset.

The user can also choose whether to save these synthetic datasets. If not, they will only be used for model training and testing before being discarded; otherwise, the newly created datasets and metadata objects will be downloaded for the user. Synthetic metrics may also be downloaded, which indicate the quality of synthetic data generated for each column.

SHAP scores were implemented with the Shap[28] library. This library provides specific implementations for certain types of models, such as tree-based models or linear models. CLASSify uses the appropriate functions for each model to provide the most efficient performance. For any ML models that do not have specific SHAP explainers, a general explainer was used. These explainers fit to the training set of the data and generate SHAP scores for each row and feature. They are averaged across all rows to produce final SHAP scores. These scores are included as part of the final report provided to the user. Because SHAP scores are not comparable between models, these scores are scaled and converted to percentages to allow for a more intuitive understanding and comparison of scores between features and models.

Visualizations are generated with matplotlib[29] and seaborn[30]. These visualizations provide comparisons of accuracy metrics between models and feature groupings through heatmaps and bar graphs, and visualizations of SHAP scores are provided as well. These visuals are stored in ClearML and presented to the user through the web interface.

The web interface included with CLASSify aims to simplify machine learning training while providing abundant training options. This interface employs a PHP[31] backend to control data flow in, out, and through the training pipelines. At a high level, a user can upload their tabular data, create a template in ClearML, and then execute a distributed training job to connected ClearML agents. However, local training is also available if a user wishes to train a model in a non-distributed fashion.

The user must first upload a training dataset, which must follow a few formatting criteria ('class' and 'index' column labels must be present, no strings other than 'TRUE' and 'FALSE'). After uploading their training dataset, a user will select from many available options, including but not limited to multiclass training, model selection, and feature evaluation. Once the desired options are selected and submitted, these options, along with the dataset, are transferred to a Python backend for error checking. The Python backend also creates a template for the training job and sends it along with the dataset to ClearML. When executing the distributed task, agents receive this template, which, until now, has yet to begin training. The web interface gives users a list of available jobs, offering options to edit, delete, or submit a job. When the Python function creating the template completes, a submit button will appear next to the dataset. Before selecting this button, a user should have one or more ClearML agents running or have a local instance prepared. Once clicked, the submit button adds the selected job to a queue defined in ClearML, which agents will enqueue and execute a job. If the agent encounters any errors, it propagates them back to the web interface, allowing the user to view and adjust them for subsequent jobs. After successful completion, the job propagates its results to the web interface, presenting images, graphs, tabular results, and SHAP scores to the user. Results from all jobs are saved in a database for future use or download until a user deletes the associated job.

Below, Images 1, 2, and 3 show screenshots of the website. On the page depicted in Image 1, users can view all of the datasets they have uploaded. If they have completed training, this is where users can view the results on those datasets. Otherwise, this is where users can access the preparation page, which is shown in Image 2. This is where users can make choices about the training, such as whether to generate synthetic data and SHAP scores, which models to train, and what parameters to use for each model. Users may choose to use default values for many of these options to ease the process, but they may also customize these values if necessary, and hovering the cursor over any of the options provides a more detailed explanation of the purpose and meaning of that option.

When viewing results, as shown in Image 3, users can see a variety of performance metrics, as well as SHAP scores for each feature. On this page, users can download any synthetic data that was generated, as well as the models that were trained, view output logs, and see visualizations.

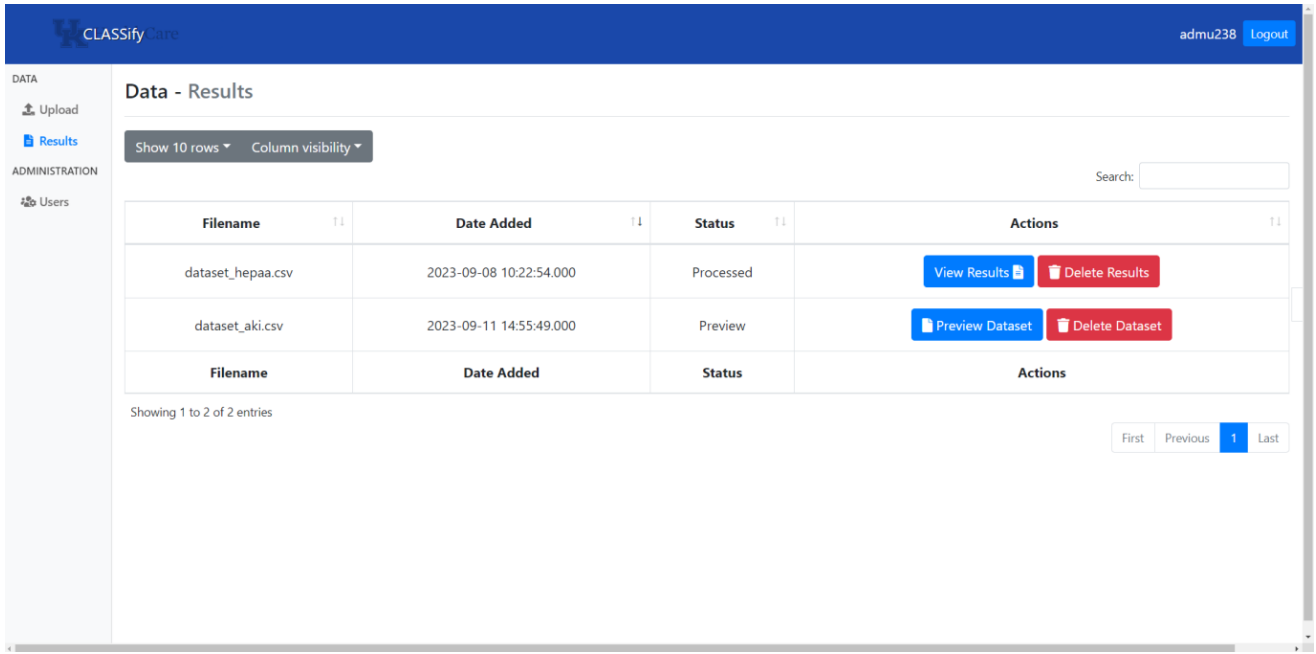

**Image 1.** Screenshot of CLASSify datasets page.

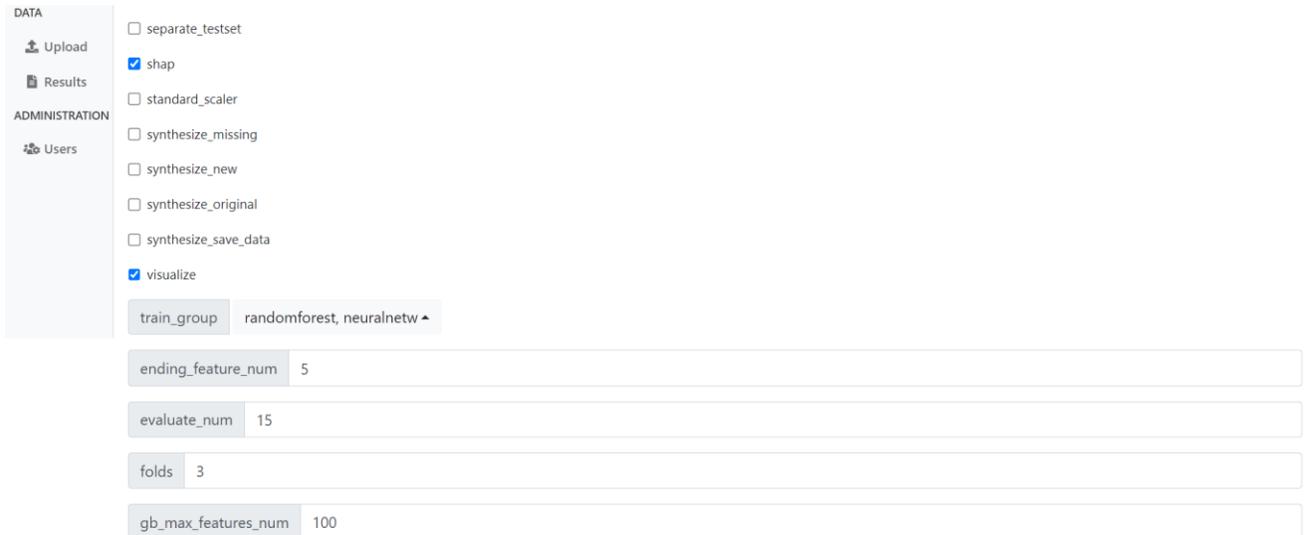

**Image 2.** Screenshot of CLASSify page where users choose options and parameters before training.

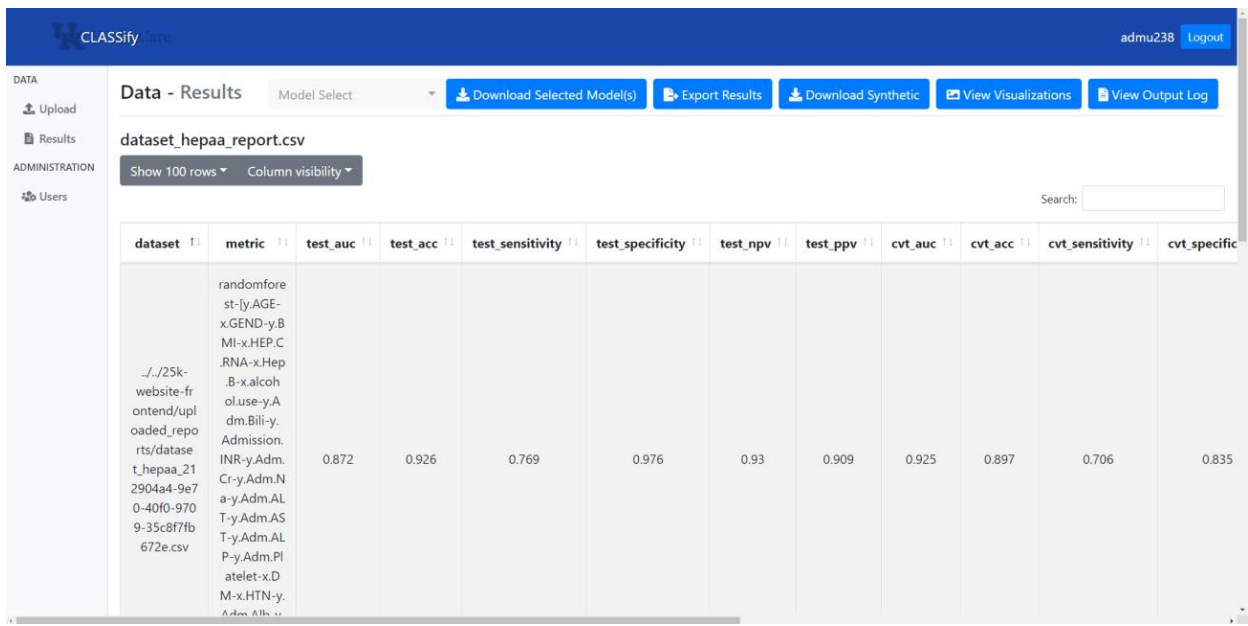

**Image 3.** Screenshot of CLASSify results page.

## Results

This section will justify the methods used in CLASSify by providing results comparisons on example datasets. While the quality of these specific results is not the focus of this paper, they are included to demonstrate the kind of metrics that are used in CLASSify and the range of different datasets and tasks that CLASSify can be used with.

All ten models were used for testing binary classification: Random Forest, Multilayer Perceptron Neural Network, XGBoost, Histogram-based Gradient Boosting, Bagging, SGD Classifier, Logistic Regression, TabPFN, Gradient Boosting, and K-Nearest Neighbors. These models were tested on ten separate binary classification datasets, five of which are internal, and five of which are publicly available. These are briefly summarized in the table below (Table 1), which provides information on the size and class distribution of each dataset.

| Dataset: | 30 Day | AKI | HEPAA | Sepsis | Soft | Breast Cancer[32] | Diabetes[33] | Heart[34] | In Out[35] | Surgery[36] |
|---|---|---|---|---|---|---|---|---|---|---|
| # Rows | 2165 | 775 | 268 | 201 | 195 | 570 | 769 | 1000 | 4000 | 14000 |
| % Positive | 50% | 17% | 24% | 46% | 49% | 37% | 35% | 13% | 40% | 25% |

**Table 1.** Size and class distribution of each tested dataset.

These datasets have a variety of lengths, feature numbers, and class balances, allowing for a full picture of how well CLASSify works on different types of datasets. While most of the results will focus on these datasets and results, multiclassification was also tested with a separate, public dataset. Referred to as Cirrhosis[37], this dataset has only 418 rows, many of which have missing data, with four separate possible classes, which are distributed 5%, 22%, 35%, and 38%. This dataset could only be tested with the four multiclass-supported models.

The datasets were tested in three different circumstances. First, parameter tuning was disabled, and each model was tested on the default parameters. Then, parameter tuning was enabled, and each model was tested again for each dataset. Finally, each dataset with synthetically balanced, and again tested with each model. Nine of the eleven total dataset results were improved with parameter tuning for 100 iterations, and every dataset with a class disparity greater than 40%/60% had improved results when synthetic balancing was implemented.

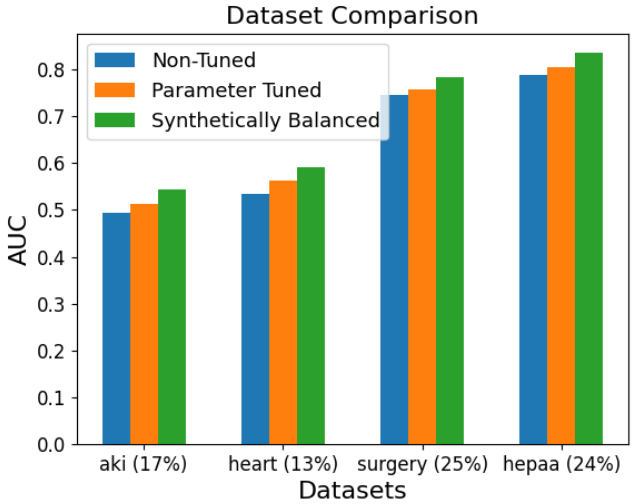

**Figure 1.** AUC comparison for each dataset, with the results of each model averaged together.

The best of these results are shown in more detail in Figure 1. This graph uses the Area Under the ROC Curve (AUC) to measure the usefulness of the model, and the results for each model are averaged together for each dataset. The percentage value given next to each dataset name represents the percentage of positive class values. Therefore, a lower percentage represents a more imbalanced dataset.

A more detailed look at each specific model's performance is given in the tables below (Tables 2 and 3). The table on the left shows the average performance of each model when parameter tuned with Optuna. The table on the right shows the best results on each example dataset.

| Model: | Average AUC: |
| --- | --- |
| Gradient Boosting | 0.727 |
| Bagging | 0.713 |
| Logistic Regression | 0.712 |
| Random Forest | 0.711 |
| Histogram Gradient Boosting | 0.701 |
| TabPFN | 0.701 |
| XGBoost | 0.699 |
| Neural Network | 0.695 |
| K-Nearest Neighbors | 0.668 |
| SGD Classifier | 0.641 |

| Dataset: | Best Model: | Best Performance: |
| --- | --- | --- |
| Breast Cancer | TabPFN | 0.988 |
| HEPAA | Random Forest | 0.872 |
| Surgery | XGBoost | 0.843 |
| Cirrhosis | Neural Network | 0.781 |
| Sepsis | K-Nearest Neighbors | 0.778 |
| Diabetes | Neural Network | 0.751 |
| In Out | Hist. Gradient Boosting | 0.747 |
| Soft | Gradient Boosting | 0.695 |
| Heart | Logistic Regression | 0.688 |
| 30 Day | Random Forest | 0.647 |
| AKI | Logistic Regression | 0.563 |

**Tables 2 and 3.** Summary of model results.

The results depend heavily on the dataset, as some are naturally easier to predict than others. But in general, gradient boosting and bagging were found to be the best-performing models, while the SGD Classifier and K-Nearest Neighbors were the worst-performing.

This same process was repeated with each dataset after it was synthetically balanced. As was discussed earlier, this had little impact on the datasets that were already nearly balanced. The significantly imbalanced datasets saw only positive changes in performance once synthetic class balancing was implemented.

A clearer comparison of different performance metrics between an original dataset and a synthetically balanced dataset is shown below in Figure 2. This graph represents the average results of each model on the HEPAA dataset to showcase the differences in metrics when the original dataset is imbalanced.

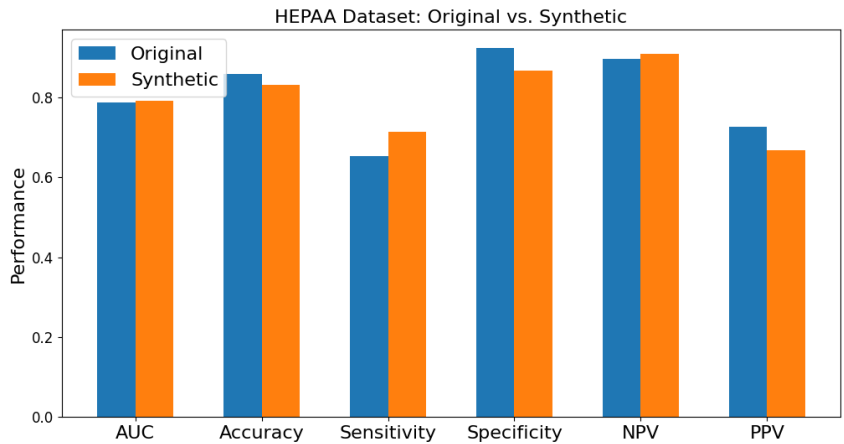

**Figure 2.** Comparison of performance metrics between original dataset and synthetically balanced dataset.

This graph demonstrates that the results of the synthetically balanced dataset are more consistent than those of the original dataset. While specificity and PPV may decrease, sensitivity and NPV are increased, meaning the model is balancing its positive and negative predictions better. However, the importance of the tradeoff between specificity and sensitivity depends on the specific application and field, so CLASSify should be fine-tuned and interpreted appropriately for any possible case.

Several metrics can be used to determine the quality of synthetic data. One is the measurement of column shapes, which gives a percentage to represent how similar the distributions of values in columns are between original and synthetic data. A higher score for a column indicates that the distribution of values in that column matches the distribution in the original dataset. An overall score for the dataset averages these values for all columns.

Another measurement is the column pair trends, which measures how well synthetic data captures the relationships and correlations between separate columns in the data. Finally, SDV provides an overall quality score to indicate how similar the synthetic data is to the original. The results for generating a new, completely synthetic dataset (AKI dataset) are shown in Figure 3. These metrics are compared for each of the four synthetic generation models (Tabular, CTGAN, Copulagan, and TVAE). Figure 3 only shows the results on a single dataset, but further analysis has shown that these trends are consistent across all datasets. This shows that the Tabular model is the most successful at capturing the trends and relationships of the original data.

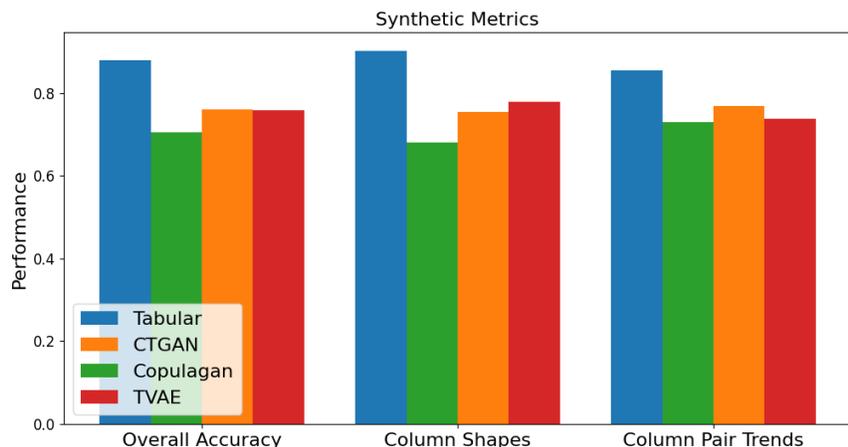

**Figure 3.** Synthetic data quality metrics.

Additionally, each fully synthetic dataset was used to train each classification model, which was tested on each original dataset to determine how useful each synthetic dataset was for training a model to predict real values. These results were consistent with the accuracy scores given above. The tabular synthesizer model's datasets performed the best, while the CTGAN and CopulaGAN synthetic datasets were not as useful in predicting real values.

There is an additional option in CLASSify to only synthesize missing values in the data. This uses the "fancyimpute"[24] module and is independent of the other synthesizer models, as those models did not have native support for imputing missing values. To test the efficacy of this feature, results were compared between the AUC of each model on an original dataset with a predefined test set, and a dataset where 20% of the feature values were randomly dropped and synthetically filled, evaluated on the same test set. In general, performance was the same or slightly worse when the dataset was synthetically filled. For example, on the HEPAA dataset, the average AUC for the original dataset, across all models, was 0.794, while the average AUC for the same dataset when 20% of the values were synthetically filled was 0.786. This shows that, while having a complete dataset is certainly preferable, CLASSify can function almost just as well with missing data.

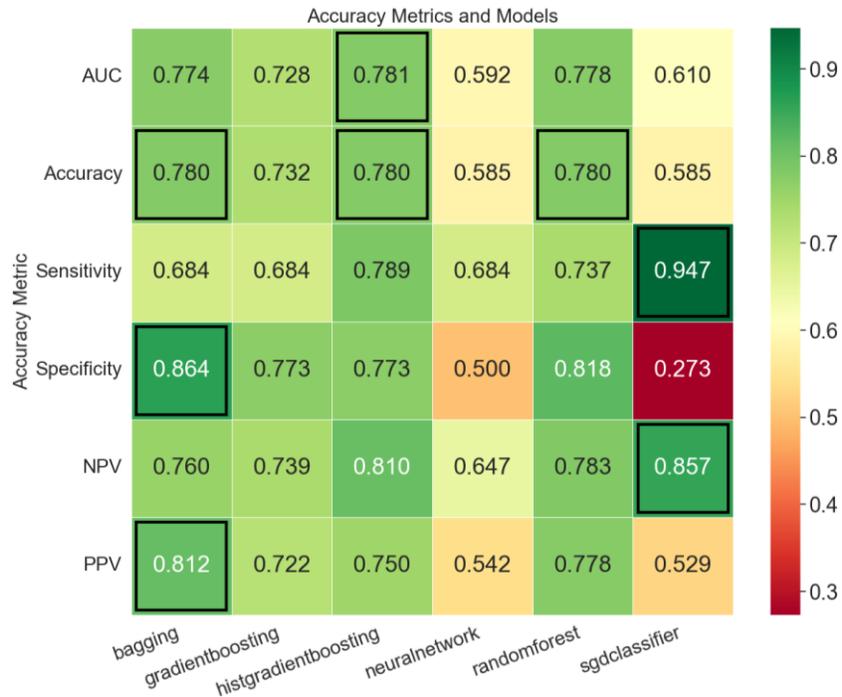

**Figure 4.** Example of provided visualization.

In addition to the final report containing performance metrics and SHAP scores for each model, several visualizations are also produced. One example of a custom plot is shown here (Figure 4). This is a heatmap representing different models and performance metrics for a given dataset, identifying the best-performing models in each area.

Additionally, the SHAP library provides visualizations for the SHAP scores. CLASSify saves the SHAP bee swarm plot, depicting several features and their positive/negative influence on the class label predictions. Each dot represents a separate row of data, and their positions indicate how much they contribute to the class prediction. As an example of how to interpret the plot shown in Figure 5, the highest feature, Weight, has the most impact on class predictions for the 30-Day example dataset. Because the positive class instances are shown to have lower weight values, and the negative class instances have higher weight values, this indicates that there is a negative correlation between the weight feature and the class label prediction.

With multiclassification, there are fewer visualization options available because certain metrics, such as specificity, sensitivity, NPV, and PPV are not calculated. However, overall comparisons of the models, as well as SHAP beeswarm plots, are still available.

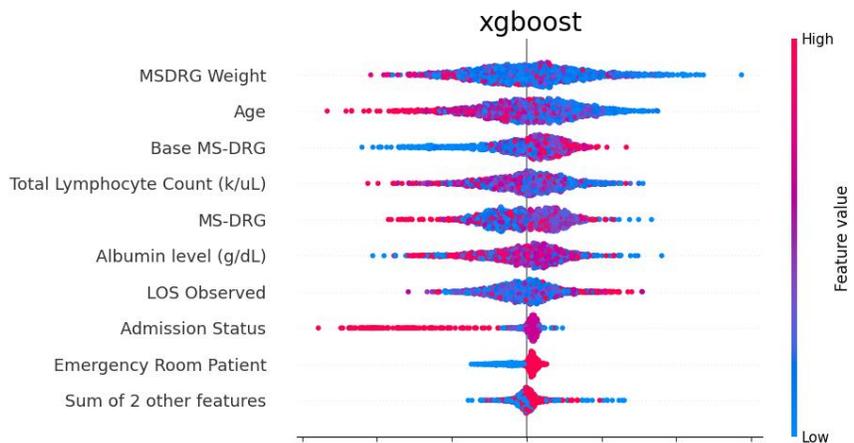

**Figure 5.** Beeswarm plot depicting SHAP scores for each feature of the example dataset.

### Discussion

The results on the test datasets show that different models perform well under different circumstances. In general, some of the best-performing models are the random forest, gradient boosting, and bagging classifiers. However, because CLASSify provides the option to customize which models the user trains, it ensures that users can only focus on the models most relevant to their problem.

There are other factors than just performance that may also influence which models are the best picks for different situations. TabPFN, due to the complex methods used in its implementation, is not made for datasets with much more than 1,000 samples in their training sets. Additionally, if time is a factor, then users may want to forgo training the Multilayer Perceptron Neural Network, which takes the longest to train and parameter tune.

CLASSify also provides the ability to build multiple versions of a given model based on different feature combinations. This is another way, apart from the SHAP scores, to evaluate the usefulness of different features and optimize the performance of the model.

Synthetic data generation is also an important aspect of CLASSify, as especially in the medical field, there is a large need for synthetic data to respect patient privacy. CLASSify allows users to save any generated data for use in other areas. Synthetic class balancing was found to have a positive effect on model performance when the original class labels were significantly imbalanced, and synthetically filling missing values was found to not have a significant effect on model performance.

One warning about synthetic data generation involves the requirement of metadata for the given dataset. This metadata describes the data types of each column in the form of a JSON. If a metadata file is not provided to CLASSify, it will generate its own file to use by parsing the data itself. However, this process can make mistakes. For example, if a particular column contains categorical data represented as numbers (i.e., integers 1 through 5 each represent a category), this column will be interpreted as numerical integers rather than categories. In practice, this was found to sometimes have negative effects on both the quality of synthetic data and the accuracy of predictions made from the synthetic data. Therefore, it is best, whenever possible, to explicitly provide a metadata file to ensure that no columns are misinterpreted.

**Conclusion**

These results show that CLASSify is a valuable tool for evaluating classification problems, generating synthetic data, and drawing explainable conclusions about data features. CLASSify provides a platform for performing complex machine learning tasks without any advanced knowledge requirement. Those experienced in the field will be able to customize complex parameters, while those without machine learning knowledge can use default settings and rely on CLASSify's automated parameter tuning. CLASSify provides many ways to compare the performances of different models and feature combinations through tables and visualizations. It also allows the user to gain a better understanding of the data itself by providing SHAP scores to indicate the relevance of different features.

While CLASSify was built and tested with the intention of use in a medical context, any tabular data can be used, so a variety of applications for this technology can be imagined. Classification problems are common in a variety of fields and CLASSify can handle generalized data. With its user-friendly interface and powerful backend, CLASSify has the ability to transform complex machine learning requirements into dependable solutions.

**References**

1. Chen X, Wang M, Zhang H. The use of classification trees for bioinformatics. WIREs Data Mining and Knowledge Discovery. 2011;1(1):55–63. doi:10.1002/widm.14
2. Kononenko I. Machine learning for medical diagnosis: History, state of the art and perspective. Artificial Intelligence in Medicine. 2001;23(1):89–109. doi:10.1016/s0933-3657(01)00077-x
3. Komura D, Ishikawa S. Machine learning approaches for pathologic diagnosis. Virchows Archive. 2019;475(2):131–8. doi:10.1007/s00428-019-02594-w
4. Dahmen J, Cook D. SynSys: A Synthetic Data Generation System for Healthcare Applications. Sensors. 2019;19(5):1181. doi:10.3390/s19051181
5. Hernandez M, Epelde G, Alberdi A, Cilla R, Rankin D. Synthetic Data Generation for Tabular Health Records: A systematic review. Neurocomputing. 2022;493:28–45. doi:10.1016/j.neucom.2022.04.053
6. Štrumbelj E, Kononenko I. Explaining prediction models and individual predictions with feature contributions. Knowledge and Information Systems. 2013;41(3):647–65. doi:10.1007/s10115-013-0679-x
7. Carney M, Webster B, Alvarado I, Phillips K, Howell N, Griffith J, et al. Teachable machine: Approachable web-based tool for Exploring Machine Learning Classification. Extended Abstracts of the 2020 CHI Conference on Human Factors in Computing Systems. 2020; doi:10.1145/3334480.3382839
8. 1. Rene L, Mario M, Laura O, Avantika L, Hernan DL. Bench-ML: A benchmarking web interface for machine learning methods and models in genomics. 2023; doi:10.1101/2023.06.05.543750
9. The Continuous Machine Learning Company [Internet]. 2023 [cited 2023 Jul 12]. Available from: https://clear.ml/
10. Learn [Internet]. [cited 2023 Jul 12]. Available from: https://scikit-learn.org/stable/
11. Breiman L. Random Forests. Machine Learning. 2001 Oct;45:5–32. doi:10.1023/A:1010933404324
12. Friedman JH. Stochastic gradient boosting. Computational Statistics & Data Analysis. 2002;38(4):367–78. doi:10.1016/s0167-9473(01)00065-2

13. 1. Chen T, Guestrin C. XGBoost. Proceedings of the 22nd ACM SIGKDD International Conference on Knowledge Discovery and Data Mining. 2016; doi:10.1145/2939672.2939785 Guryanov A. Histogram-based algorithm for building gradient boosting ensembles of piecewise linear decision trees. Lecture Notes in Computer Science. 2019;39–50. doi:10.1007/978-3-030-37334-4_4
14. Breiman L. Bagging predictors. Machine Learning. 1996;24(2):123–40. doi:10.1007/bf00058655
15. Hinton GE. Connectionist Learning Procedures. Artificial Intelligence. 1989;40(1–3):185–234. doi:10.1016/0004-3702(89)90049-0
16. Hollman N, Müller S, Eggensperger K, Hutter F. TabPFN: A Transformer That Solves Small Tabular Classification Problems in a Second. 2022; doi:10.48550/arXiv.2207.01848
17. Amari S. Backpropagation and stochastic gradient descent method. Neurocomputing. 1993;5(4–5):185–96. doi:10.1016/0925-2312(93)90006-o
18. Introduction to the logistic regression model. Applied Logistic Regression. 2005;1–30. doi:10.1002/0471722146.ch1
19. Peterson L. K-Nearest Neighbor. Scholarpedia. 2009;4(2):1883. doi:10.4249/scholarpedia.1883
20. Python API reference [Internet]. [cited 2023 Jul 12]. Available from: https://xgboost.readthedocs.io/en/stable/python/python_api.html
21. Tabpfn [Internet]. [cited 2023 Jul 12]. Available from: https://pypi.org/project/tabpfn/
22. A hyperparameter optimization framework [Internet]. [cited 2023 Jul 28]. Available from: https://optuna.org/
23. Mazumder R, Hastie T, Tibshirani R. Spectral Regularization Algorithms for Learning Large Incomplete Matrices. Journal of Machine Learning Research. 2010 Jul 9;11.
24. Fancyimpute [Internet]. [cited 2023 Jul 12]. Available from: https://pypi.org/project/fancyimpute/
25. Welcome to the SDV! [Internet]. [cited 2023 Jul 12]. Available from: https://docs.sdv.dev/sdv/
26. Fast ML preset [Internet]. [cited 2023 Jul 12]. Available from: https://docs.sdv.dev/sdv/single-table-data/modeling/synthesizers/fast-ml-preset
27. Xu L, Skoularidou M, Cuesta-Infante A, Veeramachaneni K. Modeling tabular data using conditional gan [Internet]. 2019 [cited 2023 Jul 12]. Available from: https://arxiv.org/abs/1907.00503
28. Welcome to the shap documentation [Internet]. [cited 2023 Jul 12]. Available from: https://shap.readthedocs.io/en/latest/
29. Visualization with python [Internet]. [cited 2023 Jul 12]. Available from: https://matplotlib.org/
30. Statistical Data Visualization [Internet]. [cited 2023 Jul 17]. Available from: https://seaborn.pydata.org/
31. Bakken SS, Suraski Z, Schmid E. PHP Manual: Volume 1. iUniverse, Incorporated; 2000.
32. Mansy M. Breast cancer dataset [Internet]. 2023 [cited 2023 Jul 26]. Available from: https://www.kaggle.com/datasets/mahmoudelmansy/breast-cancer-dataset
33. Rajendran S. Diabetes prediction [Internet]. 2022 [cited 2023 Jul 26]. Available from: https://www.kaggle.com/datasets/cpluzshrijayan/diabetes-prediction
34. Murattademir. Heart disease - binary classification [Internet]. Kaggle; 2022 [cited 2023 Jul 26]. Available from: https://www.kaggle.com/code/murattademir/heart-disease-binary-classification
35. Sadikin M. EHR dataset for Patient Treatment Classification [Internet]. Mendeley Data; 2020 [cited 2023 Jul 26]. Available from: https://data.mendeley.com/datasets/7kv3rctx7m/1
36. M D. Dataset surgical binary classification [Internet]. 2018 [cited 2023 Jul 26]. Available from: https://www.kaggle.com/datasets/omnamahshivai/surgical-dataset-binary-classification
37. 1. Fedesoriano. Cirrhosis prediction dataset [Internet]. 2021 [cited 2023 Aug 25]. Available from: https://www.kaggle.com/datasets/fedesoriano/cirrhosis-prediction-dataset